\documentclass{article}
\usepackage{spconf,amsmath,graphicx}
\usepackage{enumitem}
\setlist{nosep, leftmargin=14pt}
\usepackage{mwe} 
\usepackage{multirow}
\usepackage{makecell}

\title{Low performing pixel correction in computed tomography with unrolled network and synthetic data training}
\name{Hongxu Yang, Levente Lippenszky, Edina Timko, Lehel Ferenczi, Gopal Avinash  \thanks{Contact author: hongxu.yang@gehealthcare.com}}
\address{STO-Artificial Intelligence \& Machine Learning, GE HealthCare}
\begin{document}
%
\maketitle
\begin{abstract}
Low performance pixels (LPP) in Computed Tomography (CT) detectors would lead to ring and streak artifacts in the reconstructed images, making them clinically unusable. In recent years, several solutions have been proposed to correct LPP artifacts, either in the image domain or in the sinogram domain using supervised deep learning methods. However, these methods require dedicated datasets for training, which are expensive to collect. Moreover, existing approaches focus solely either on image-space or sinogram-space correction, ignoring the intrinsic correlations from the forward operation of the CT geometry. In this work, we propose an unrolled dual-domain method based on synthetic data to correct LPP artifacts. Specifically, the intrinsic correlations of LPP between the sinogram and image domains are leveraged through synthetic data generated from natural images, enabling the trained model to correct artifacts without requiring any real-world clinical data. In experiments simulating 1-2\% detectors defect near the isocenter, the proposed method outperformed the state-of-the-art approaches by a large margin. The results indicate that our solution can correct LPP artifacts without the cost of data collection for model training, and it is adaptable to different scanner settings for software-based applications.  
\end{abstract}
\section{Introduction}
\label{sec:intro}
\begin{figure}[ht!]
\centering{\includegraphics[width=8cm]{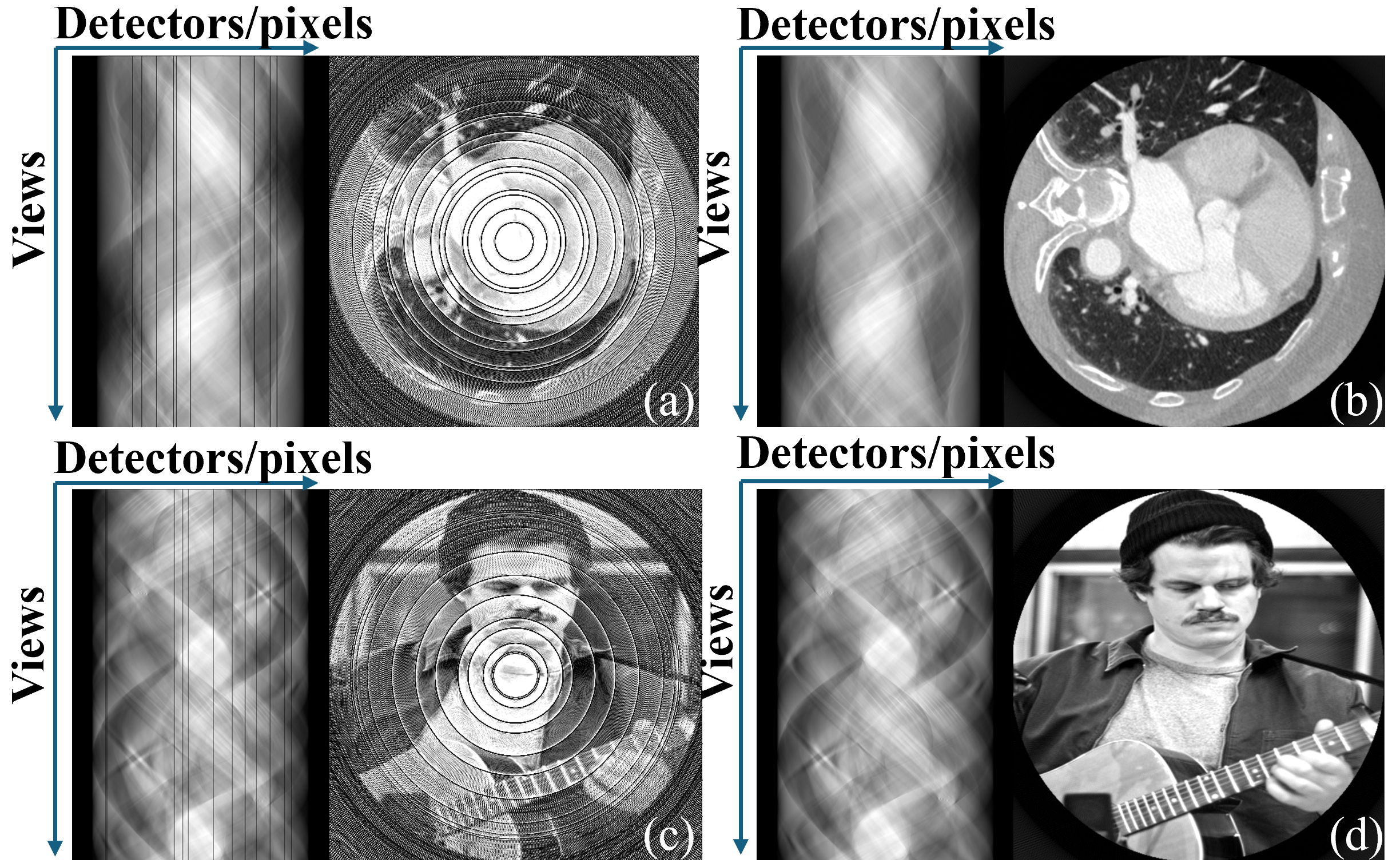}}
\caption{Example CT and synthetic images with artifacts. Top: sinogram and corresponding CT with and without LPP. Bottom: sinogram and corresponding synthetic image with and without LPP. LPP: Zero-out lines along view axis.}
\label{img_compare}
\end{figure}

Computed Tomography (CT) is an important imaging technique in clinical workflows, providing detailed visualization of different patient anatomies, assisting disease diagnosis and intervention guidance. These qualities make CT critically important in medical imaging, which requires high image quality with low (or no) artifacts. Nevertheless, due to inherent physical hardware limitations, artifacts due to detector imperfection can be a crucial issue, which requires careful correction to achieve high image quality. Defective detector pixels can lead to missing signals in the sinogram of the measurement and are commonly defined as low performing pixels (LPP). The most common artifacts caused by LPP are ring and streak artifacts. The severity of artifacts depends on the LPP position; the closer an LPP is to the isocenter, the greater the image degradation, which may lead to an unusable image~\cite{patil2022deep}, examples are shown in Fig.~\ref{img_compare}. Correcting this LPP artifact requires full detector module replacement, which comes with high service cost, and prolongs waiting times for users. 

Recent studies have proposed different solutions to correct LPP generated artifacts by using software solutions, such as image processing techniques or deep learning (DL) methods. Specifically, interpolation-based techniques were commonly used before the era of deep learning and achieved reasonable performance~\cite{abdoli2011reduction,salehjahromi2019directional}. More recently, several DL methods have been developed to address LPP artifacts in the image or sinogram domain. DeepRAR~\cite{trapp2022deeprar} was proposed to correct the ring artifact in CT image using UNet~\cite{cciccek20163d}. Other image-based DL models were also considered for image restoration, such as CNN-based NAFNet~\cite{chen2022simple, wu2024unsupervised} and transformer-based AST~\cite{zhou2024adapt}. Alternatively, correction DL models in sinogram were also proposed, such as Conjugate-CNN~\cite{patil2022deep} with supervised learning, SinoRAR~\cite{shi2025ring} and Riner~\cite{wu2024unsupervised} using implicit neural representation method (INR). Most deep learning methods have shown promising performance, but still suffer from common limitations of supervised learning: high data collection costs and limited generalization. Although INR-based methods can alleviate data requirements for model training, their high computational cost in test-time optimizations remains a key bottleneck for real-world applications. 

In this paper, we propose a novel LPP correction method. Unlike conventional supervised models trained on real medical data, we propose to use synthetic data integrate root cause of the LPP in both detector and image domain. With the generated training pairs under supervised learning, the LPP artifacts are corrected by a dual-domain unrolled method, which iteratively corrects the artifacts in detection and image spaces. Our main contributions are threefold. First, a natural image-based synthetic data generation strategy for LPP artifact, which generates unbiased training pairs. Second, we correct the LPP by a dual-domain unrolled model, which reformulates the defective measurement errors as a compressed sensing problem in the data. Finally, experiments on a public CT dataset demonstrate that our method outperforms existing approaches trained on real CT data by a large margin.

\begin{figure}[ht!]
\centering{\includegraphics[width=8cm]{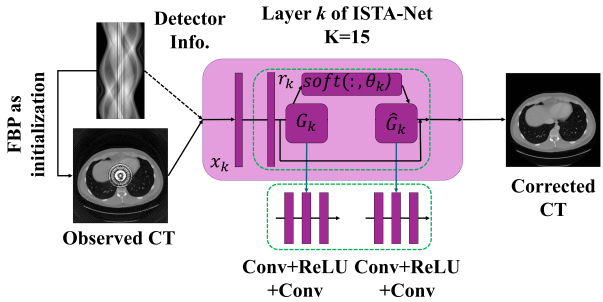}}
\caption{Structure of ISTA-Net. The observed sinogram with LPP is reconstructed by FBP to generate observed CT, which is processed by k-layer ISTA-Net. In addition to the CT image, the detector LPP information is also provided as input to ISTA-Net characterizing the sinogram defect pattern. Output is the corrected CT without LPP artifacts.}
\label{model_design}
\end{figure}

\section{Method}
\subsection{Theoretical Analysis of LPP Artifacts}
A standard CT detector works as an energy-integrating detector. In a CT scanner where the source-detector pair rotates around the gantry by $360^\circ$, the information received by the detector channels corresponds to specific scanning views~\cite{patil2022deep}. For a given view, the readout from each detector channel is produced by the line integral of the X-ray attenuation through the materials along the path of an X-ray beam with energy $E$. Based on Lambert-Beer's law, the readout data is defined as 
\begin{equation}
I(E)=N\eta(E)\exp(-\int_{0}^{l_1}\mu(E,\vec{l})dl)
\end{equation}
where $N$ is the total number of incidental photons, $\eta(E)$ is the detector response function w.r.t. photon energy, $\mu(E,\vec{l})$ is the energy attenuation coefficient at energy level $E$, vector $\vec{l}$ is location and ${l_1}$ is the propagation path length. 

As for LPP, the received signal can vary due to differences in the response functions. When a detector pixel fails or reaches a near-dead state, the response function $\eta(E)$ is approaching zero. Unlike completely dead pixels, the open circuit may generate any unstable value due to diode variations potentially affecting neighborhood detectors. In this paper, we focus on the former case, where the detectors are entirely dead and do not produce a signal. As shown in Fig.~\ref{img_compare}, a single LPP in the sinogram data shows complete missing data in a specific channel. The corresponding image reconstructed by filtered back projection (FBP) includes ring and streak artifacts when no correction is applied. Since LPPs do not respond to X-ray energy, they can be identified in the measured data by locating regions of extremely low signal in the sinogram. As LPP detection is out of scope in this paper, the LPP location is assumed to be known prior to correction.

\subsection{Problem Formulation}
Existing methods correct artifacts in the image domain~\cite{trapp2022deeprar} as standard restoration tasks, or correct the missing data in the sinogram domain as an interpolation task~\cite{patil2022deep,shi2025ring}. Although Riner~\cite{wu2024unsupervised} proposed a joint dual-domain method to correct the artifacts, its test time optimization is time-consuming and requires intensive calculations for each grid point. In our proposed method, we consider the LPP in the sinogram domain as a compressed sensing problem, where LPP is formulated by randomly generated masks that is used to undersample data points from the perfect raw data measured at the detector channels. Mathematically, the image correction problem can be formulated as 
\begin{equation}
\arg \min_{x} \frac{1}{2}||Ax - y||_2^2 + \lambda R(x)
\end{equation}
where $x$ is the desired image, $y$ is the observed sinogram data with LPP, $A$ is the forward projection, such as Fan-beam, Parallel-beam or Cone-beam geometry, together with known position of the failure detectors. The function $R(x)$ denotes the regularization with the weight parameter $\lambda$. The purpose of CT artifact correction is to recover the desired image $x$ from its measurement $y$. 

To solve the above function, we used ISTA-Net as the solver, which is an unrolled version of the Iterative Shrinkage Thresholding Algorithm (ISTA) via CNN~\cite{zhang2018ista}, as demonstrated in Fig.~\ref{model_design}. Specifically, ISTA-Net solves the image correction problem by iteratively applying two steps:
\begin{equation}
r^{(k)}=x^{(k-1)} - \rho A^{-1}(Ax^{(k-1)} - y)
\end{equation}
\begin{equation}
x^{(k)}=\hat{G}(soft(G(r^{(k)}), \theta))
\end{equation}
where $k$ denotes the iteration index, $\rho$ is the step size, $G(\cdot)$ and $\hat{G}(\cdot)$ are CNN-based image transformations, with $\hat{\cdot}$ denoting the symmetrical operation of the structure. Parameter $\theta$ represents the shrinkage threshold for the soft thresholding operation $soft$. Operation $A$ is the forward projection of the scanning protocol, while $A^{-1}$ is the reverse of the forward projection, commonly known as FBP. All parameters in the ISTA-Net can be optimized by end-to-end training~\cite{zhang2018ista}.

\subsection{Synthetic Data for Training}
Commonly used image restoration approaches, such as CNN- or Transformer-based solutions, require image and ground truth (GT) pairs for training, which introduce data collection efforts and generalization limitation. In addition, medical data collected under specific region-of-interest and scanner settings cannot be transferred to a different scanner protocol, creating additional requirements for data collection across different body parts, vendors and scanners. With the above formulation of LPP, the DL model will focus only on data filling with the guidance of transformation $A$ in CT. Therefore, the model is expected to correct the artifacts in the dual-domain iteratively without considering the actual content of the image from any resources.

In our settings, we obtained our natural image dataset from generic datasets by converting the RGB images to grayscale. For a grayscale image, the maximum intensity is randomly scaled to a linear attenuation coefficient $\mu$ within the empirical range $[0.5, 0.7]$. Then, with known predefined CT geometry, such as a Fan-beam CT in this paper, the forward operation is applied to the converted grayscale image to obtain the sinogram data. To simulate the LPP on the detector, a random number of detector signals are zeroed out, which are shown in Fig.~\ref{img_compare}. Finally, the corrupted sinogram is reconstructed as synthetic CT using a standard filtered back-projection algorithm with Ramp filter. With the synthetic CT image with known LPP on simulated detectors, the correction model is forced to learn artifact correction function regardless of the actual real context of the signal. Therefore, it is expected to have a more generalized performance. 
\section{Experiments and results}
\subsection{Dataset}
The ISTA-Net for artifact removal was trained on synthetic images designed to mimic CT characteristics. This dataset was generated using approximately 5k color images from the public ILSVRC2017 dataset~\cite{ILSVRC15}. These images were converted to grayscale and resized to $512\times512$ pixels and then transformed into the sinogram space with a predefined geometry based on TorchRadon~\cite{ronchetti2020torchradon} parameters. Specifically, we defined a Fan-beam CT scanner with 681 detectors for 984 views over a full $2\pi$ rotation. The distances from the rotation center to both the source and the detector were set to 722 units. To simulate LPP, we have randomly selected 7 to 15 detectors within detector index range of $[80, 601]$, which were zeroed-out acting as failure detectors. 

In addition to the synthetic dataset, we evaluated the proposed unrolled LPP correction method trained using the MMWHS CT dataset~\cite{zhuang2018multivariate}. Specifically, 20 volumes of training images were used to generate CT images with LPP artifacts for comparison with other supervised learning methods, while 40 volumes were held out for testing and validation. For each training volume, axial slices containing semantic masks from MMWHS~\cite{zhuang2018multivariate} were extracted for training without restriction, while the central axial slice for each testing volume was used as the evaluation image. This process resulted in around 5k CT slices for training and 40 CT slices for testing. The HU values are transformed to $\mu$ unit by using $\mu_{water}=0.268$ and $\mu_{air}=0$. The LPP simulation and Fan-beam geometry followed the synthetic data generation strategy described in Section 2.3. To further evaluate the generalization for background context, additional 80 testing CT images are obtained from RibFrac dataset~\cite{yang2025deep} with same generation strategy.

\begin{figure*}[ht!]
\centering{\includegraphics[width=16cm]{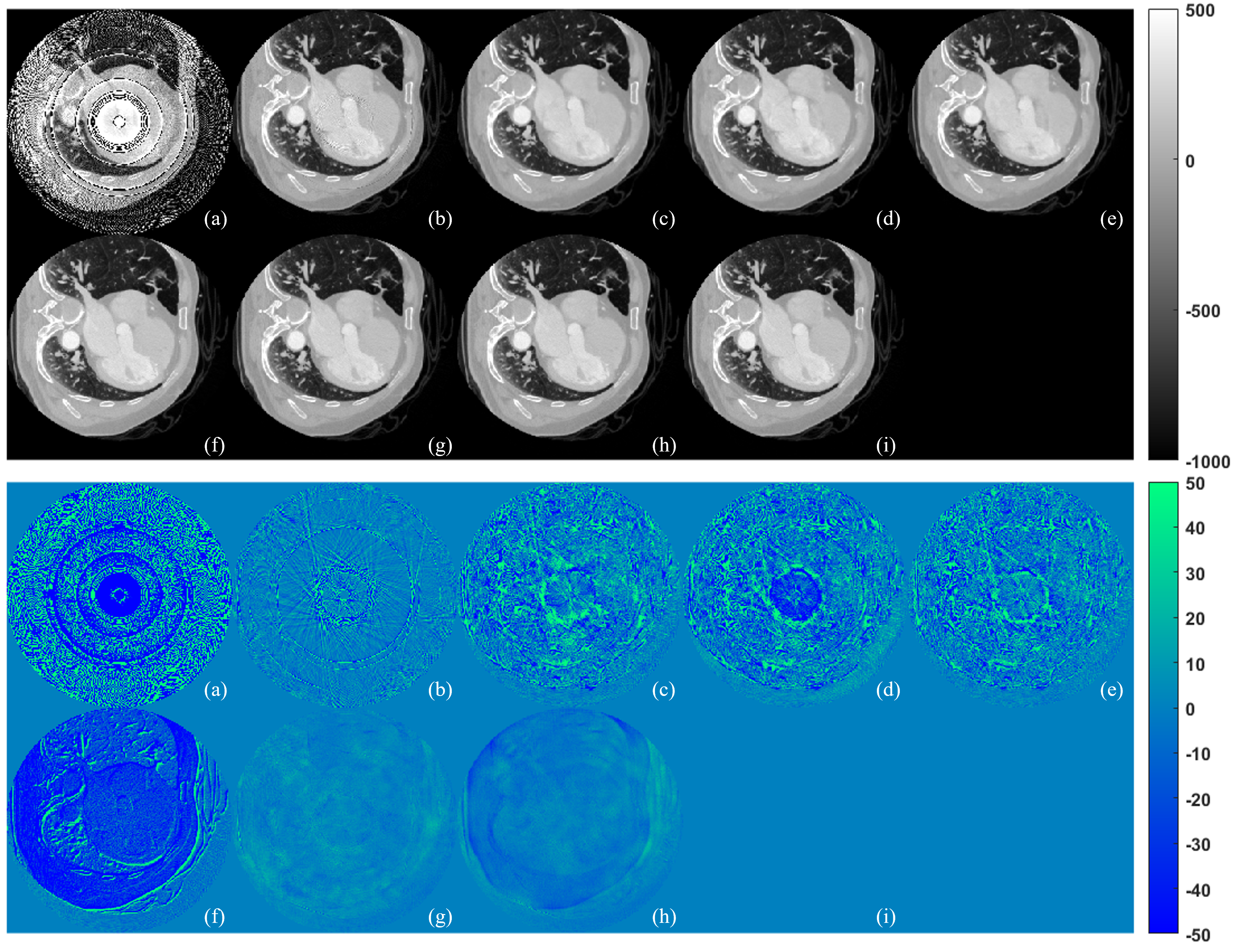}}
\caption{Qualitative comparison for LPP artifact removal methods (a) FBP, (b) Interpolation, (c) AST, (d) DeepRAR, (e) NAFNet, (f) Riner, (g) Ours-Real (h) Ours-Syn, and (i) Ground truth. Top: images in HU value. Bottom: HU value differences to GT.}
\label{results_compare}
\end{figure*}

\begin{table}[htbp]
\caption{Performance comparisons of correction results by Mean Absolute Error (MAE) in HU, Peak Signal-to-Noise Ratio (PSNR) in dB and Structural Similarity Index Measure (SSIM) on testing images, measured by mean (std.).}
\label{performance}
\begin{tabular}{lccc}
\hline
Method           & MAE $\downarrow$ & SSIM $\uparrow$ & PSNR $\uparrow$\\ \hline
FBP              &   $\sim$200       &    0.327(0.062)       &  13.9(1.7)    \\ 
Inter.    &      6.2(3.1)    &      0.906(0.075)     &   41.7(4.0)   \\ \hline
AST              &    13.8(3.1)      &    0.772(0.057)       &   40.0(2.0)   \\ 
DeepRAR          &     18.9(10.9)     &      0.736(0.068)     &    38.2(2.6)  \\ 
NAFNet           &    13.5(3.1)      &     0.771(0.058)      &   40.4(2.1)   \\   
Riner            &      29.7(4.1)    &     0.784(0.042)      &    33.8(1.0)  \\ \hline
Ours (Real)        &    2.8(0.4)      &     0.985(0.006)      &   54.1(1.3)   \\ 
Ours (Syn) &      4.2(0.7)    &    0.964(0.018)       &   50.2(1.7)   \\ \hline
Inter.    &      8.1(4.3)    &      0.872(0.097)     &   41.0(3.0)   \\ 
Ours (Real/Rib)        &    9.2(2.7)      &     0.869(0.044)      &   46.1(3.4)   \\ 
Ours (Syn/Rib) &      6.0(1.3)    &    0.924(0.027)       &   49.4(2.2)   \\ 
\hline
\end{tabular}
\end{table}

\subsection{Experiment settings}
The proposed method uses ISTA-Net as the solver, configured with a 15-layer architecture~\cite{zhang2018ista} and trained by AdamW optimizer with a learning rate of 0.0001. We compared the proposed method to several SOTA image restoration solutions, including (1) filtered-back projection (FBP), and interpolation (Inter.) methods, (2) supervised DL models, such as AST~\cite{zhou2024adapt}, DeepRAR~\cite{trapp2022deeprar}, NAFNet~\cite{chen2022simple}, on MMWHS training images, and (3) most recent unsupervised Riner model with Implicit Neural Representation (INR) solution~\cite{wu2024unsupervised}. The compared models were trained on real data ,while our method was evaluated by two cases: trained on real CT images (Real), and trained on synthetic images (Syn). The experiments were conducted on one A100 GPU trained for 100 epochs. In addition, our methods are also evaluated on RibFrac images, denoted as Rib. The results are measured by Peak Signal-to-noise Ratio (PSNR), Mean Absolute Error (MAE) and Structural Similarity Index Measure (SSIM) in HU values.

\subsection{Experiment results}
Qualitative results are shown in Fig.~\ref{results_compare}, and quantitative results are summarized in Table~\ref{performance}. From the results, the proposed method achieved the best performance for LPP artifacts correction. Interpolation achieved the second-best results, which is also much better than other image restoration solutions in image domain by DL, which implies that simple correction in the sinogram domain can provide much better improvement than complex algorithms in image. All supervised methods trained on CT images, such as AST, DeepRAR and NAFNet can correct the artifacts but do not outperform our solution. The state-of-the-art Riner with INR got the worst results, along with this being the most time-consuming model, which requires online optimization for each observed image. The proposed method achieved the highest performance in all three metrics. Although our method trained on synthetic data performed slightly worse than the fully supervised counterpart trained on real CT images, it inherently avoids anatomy-specific biases, which is shown by RibFrac (Syn/Rib) images with better results.   

\section{Conclusions}
This paper presents a novel low performance pixel correction method for computed tomography images, which considers an unrolled deep learning model using synthetic images without real clinical data. This method introduces an alternative dual-domain strategy for mitigating LPP artifacts, demonstrating superior performance compared to other solutions. Through dedicated evaluation on a simulated dataset, the proposed method showed promising results, highlighting its potential to reduce the cost of scanning maintenance via software-based in DICOM artifact correction. Future study will focus on generalization analysis of synthetic data trained model with different anatomies under different scanner setups for real-world dataset in one model.
\bibliographystyle{IEEEbib}
\bibliography{strings,refs}

\end{document}